\documentclass{article}

\usepackage{microtype}
\usepackage{graphicx}
\usepackage{tikz}
\usetikzlibrary{shapes, arrows, positioning, shadows, fit}
\usepackage{subfigure}
\usepackage{booktabs} 

\usepackage{hyperref}

\usepackage[accepted]{icml2025}

\usepackage{amsmath}
\usepackage{amssymb}
\usepackage{mathtools}
\usepackage{amsthm}

\usepackage[capitalize,noabbrev]{cleveref}

\theoremstyle{plain}
\newtheorem{theorem}{Theorem}[section]

\theoremstyle{definition}
\newtheorem{definition}[theorem]{Definition}

\theoremstyle{remark}

\usepackage[textsize=tiny]{todonotes}

\icmltitlerunning{Enhancing Memory Recall in LLMs with Gauss-Tin}

\begin{document}

\twocolumn[
    \icmltitle{Enhancing Memory Recall in LLMs with Gauss-Tin: \\
               A Hybrid Instructional and Gaussian Replay Approach}
    
    
    
    \icmlsetsymbol{equal}{*}
    
    \begin{icmlauthorlist}
    \icmlauthor{Iing Muttakhiroh}{equal,yyy}
    \icmlauthor{Thomas Fevens}{equal,yyy}

    \end{icmlauthorlist}
    
    \icmlaffiliation{yyy}{Department of Computer Science and Software Engineering, Concordia University, Montreal, Canada}
    
    \icmlcorrespondingauthor{Iing Muttakhiroh}{iing.muttakhiroh@concordia.ca}
    
    \icmlkeywords{Machine Learning, ICML, Continual Learning, LLMs, Catastrophic Forgetting }
    
    \vskip 0.3in
]



\printAffiliationsAndNotice{\icmlEqualContribution} 

\begin{abstract}
Despite the significant advancements in Large Language Models (LLMs), catastrophic forgetting remains a substantial challenge, where models lose previously acquired knowledge upon learning new information. Continual learning (CL) strategies have emerged as a potential solution to this problem, with replay-based techniques demonstrating superior performance in preserving learned knowledge. In this context, we introduce Gauss-Tin, a novel approach that integrates the replay strategy with a Gaussian mixture model to enhance the quality of sample selection during training, supplemented by instructional guidance to facilitate the generation of past learning. This method aims to improve LLMs' retention capabilities by strategically reinforcing important past learnings while accommodating new information. Our experimental results indicate a promising 6\% improvement in retention metrics over traditional methods, suggesting that Gauss-Tin is an effective strategy for mitigating catastrophic forgetting in LLMs. This study underscores the potential of hybrid models in enhancing the robustness and adaptability of LLMs in dynamic learning environments.

\end{abstract}

\section{Introduction}
\label{introduction}
Large Language Models (LLMs) have significantly impacted the machine learning industry, enabling advances in diverse applications such as natural language processing, machine translation, and AI-driven content generation \cite{radford2018improving,touvron2023llama}. Despite their success, LLMs face a critical challenge known as catastrophic forgetting. This phenomenon occurs when an LLM trained on new tasks inadvertently overwrites knowledge acquired from previous tasks. For example, an LLM trained to perform sentiment analysis might lose its proficiency in language translation if not continually trained on both tasks
\cite{yang2024recent,budnikov2024generalization}.

Continual Learning (CL) has emerged as a promising approach to mitigate this issue by allowing models to retain old knowledge while adapting to new information \cite{rolnick2019experience,kirkpatrick2017overcoming,mallya2018packnet}. Implementing CL in LLMs presents a fertile area for research, as traditional strategies like CT0, RCL, and DynaInst, while innovative, often require extensive computational resources and may not efficiently generalize across diverse tasks \cite{scialom2022fine,wang2024comprehensive,mok2023large}.Methods like CITB and SSR, although designed to combat forgetting, frequently fail to maintain performance uniformly across different tasks due to their static nature \cite{zhang2023citb,huang2024mitigating}. Similarly, approache like ConTinTin, though tailored to specific tasks, struggle with broad applicability and scalability \cite{yin2022contintin}. 

In response to these limitations, we propose Gauss-Tin,  a memory-efficient framework based on a Gaussian Mixture Model (GMM) to address catastrophic forgetting in LLMs. Gauss-Tin leverages a GMM to generate the most representative exemplars from previous tasks. This method allows us to selectively store only the most crucial examples, significantly reducing memory requirements without compromising performance. Furthermore, we enhance the selection process with task-specific prompts, aiding the GMM in identifying and generating the most effective exemplars. This approach not only improves the efficiency of memory usage but also ensures that LLMs maintain high performance across continually evolving tasks.

\section{Related Works}
This section reviews prior research in the proposed area, focusing on CL with an emphasis on replay methods. It then delves into more specific scenarios, particularly the task-incremental setting, and explores strategies for overcoming catastrophic forgetting through the use of GMMs and prompt engineering.

\subsection{Continual Learning Meets Large Language Models}
Using CL to address catastrophic forgetting in LLMs is gaining increasing attention. Unlike traditional machine learning, the learning phase in LLMs introduces unique challenges, making the application of CL in this context slightly different from traditional approaches. Broadly, CL can be categorized into vertical continual learning and horizontal continual learning \cite{shi2024continual,yang2024recent}. Our work focuses on horizontal continual learning, specifically during the continual fine-tuning stage. Several recent works have explored this area, particularly by applying replay strategies. RCL \cite{xiao2024improving} employs a reinforcement learning approach for replay selection, but its complexity makes it computationally intensive. DynInst \cite{song2023dynamics} dynamically adjusts the importance of replayed samples but is prone to instability in long sequences of tasks. CITB \cite{zhang2023citb} integrates task-specific replay buffers but often suffers from memory constraints. ConTinTin \cite{yin2022contintin} utilizes task-based prompts for replay but lacks adaptability across diverse tasks. Finally, InsCL \cite{wang2024inscl} optimizes replay through exemplar selection but has limited effectiveness when faced with highly imbalanced datasets. These limitations highlight the need for a more robust replay strategy that combines high-quality exemplar generation and efficient task adaptation. 

\subsection{Gaussian Mixture Model for Overcoming Catastrophic Forgetting}
Using probabilistic models to generate samples can create high-quality examples, but this approach is still not well-explored. Unlike generative models that just create samples, fully probabilistic models model the data distribution directly. This is crucial for CL because it helps in detecting outliers and generating samples \cite{bagus2021investigation,pfulb2021continual}. Gaussian Mixture Replay (GMR) was introduced as a rehearsal method for CL that uses GMMs. It helps prevent catastrophic forgetting by mixing samples from previous tasks into current training data. GMMs are versatile, aiding in generating samples, estimating densities for detecting outliers, and offering feature representations for classification. This approach also reduces memory use and keeps processing time steady across tasks. However, depending on GMMs might restrict its use with more complex datasets \cite{pfulb2021overcoming}. More recent work is PromptCCD \cite{cendra2025promptccd}. They utilize a GMM as a prompting method for Continual Category Discovery. At its core is the Gaussian Mixture Prompting (GMP) module, a dynamic pool that updates over time to facilitate representation learning and prevent forgetting during category discovery. This dynamic prompting not only improves exemplar utility but also enables the model to generalize better across categories. However, its application outside category discovery tasks remains limited, highlighting the need for broader generalizability. 
\cite{krawczyk2024adiabatic} introduce Adiabatic Replay (AR) to mitigate catastrophic forgetting. Traditional replay-based methods often require the rehearsal of all previously learned knowledge during each new learning phase, leading to increased computational demands as knowledge accumulates. In contrast, AR operates under the assumption that each new learning phase introduces only a small addition to existing knowledge—termed an ``adiabatic" change. By GMMs, AR selectively updates its internal representations only where data statistics have changed, thereby reducing the need for extensive replay. This selective replay focuses on samples similar to the new data, enhancing efficiency.

\subsection{Prompt Engineering for Continual Learning}
Prompt engineering has emerged as a vital technique for improving the performance of LLMs in various tasks, including CL. By designing task-specific instructions or prompts, prompt engineering enables LLMs to better align with learning objectives, facilitating the generation of high-quality exemplars and enhancing model adaptability. One prominent work is Prompt Conditioned Variational Autoencoder (PCLL) \cite{zhao2022prompt}, which leverages prompts to guide exemplar generation for lifelong learning in task-oriented dialogue systems. PCLL conditions the generative model on task-specific prompts, enabling effective replay of past knowledge while minimizing catastrophic forgetting. Despite its strengths, this approach is domain-specific and lacks generalization across broader CL scenarios. Recent advancements, such as Progressive Prompts \cite{razdaibiedina2023progressive}, introduce a method where prompts are sequentially concatenated for new tasks while keeping the base model frozen. This approach allows forward transfer and mitigates forgetting but can lead to inefficiencies as the number of tasks grows. Another work is CODA-Prompt \cite{smith2023coda}, a method designed for rehearsal-free CL. It uses a decomposed attention-based prompting mechanism where a set of prompt components is dynamically assembled based on input-conditioned weights. These works collectively illustrate the potential of prompt engineering in CL. By integrating task-specific instructions and leveraging generative replay, prompt engineering offers a pathway to enhance the effectiveness of LLMs in the CL framework. However, challenges such as scalability across diverse tasks, and computational efficiency remain open areas for exploration. Our work builds on this foundation by combining prompt engineering with Gaussian Mixture Models to further improve exemplar quality and mitigate forgetting in LLMs. 

\section{Methodology}
Our approach addresses catastrophic forgetting challenges in LLMs by applying CL techniques. Specifically, we adopt a replay strategy, in which samples from previous tasks are generated using Gauss-Tin, a framework that leverages the strength of GMMs for learning feature representations and the power of instruction-based guidance to refine the selection process. 

\subsection{Problem Formulation}
Our research revolves around a general model consisting of two pivotal components: an LM-based task solver and a GMM-based generator. These components are designed to work in synergy within a CL framework, addressing the challenges posed by sequential task learning and exemplar generation based on task-specific latent distributions.


\begin{figure}[htbp]
\centering
\resizebox{\columnwidth}{!}{%
\begin{tikzpicture}[
    node distance=2.5cm,
    mlbox/.style={rectangle, rounded corners=8pt, draw=black, line width=1.5pt, 
                  fill=blue!15, minimum width=3cm, minimum height=1.8cm, 
                  text centered, font=\sffamily\bfseries, align=center},
    databox/.style={rectangle, rounded corners=8pt, draw=orange!80, line width=1.5pt,
                    fill=orange!20, minimum width=3cm, minimum height=1.8cm,
                    text centered, font=\sffamily\bfseries, align=center},
    processbox/.style={rectangle, rounded corners=8pt, draw=green!80, line width=1.5pt,
                       fill=green!20, minimum width=3cm, minimum height=1.8cm,
                       text centered, font=\sffamily\bfseries, align=center},
    modelbox/.style={rectangle, rounded corners=8pt, draw=purple!80, line width=1.5pt,
                     fill=purple!20, minimum width=3cm, minimum height=1.8cm,
                     text centered, font=\sffamily\bfseries, align=center},
    containerbox/.style={rectangle, draw=black, line width=1.5pt,
                     minimum width=3cm, minimum height=1.8cm,
                     text centered, font=\sffamily\bfseries, align=center},
    mlarrow/.style={->, line width=2pt, color=gray!70, >=stealth,font=\sffamily\small},
    feedbackarrow/.style={->, line width=2pt, color=red!60, >=stealth, dashed,font=\sffamily\small},
    merge/.style={circle, draw=green!80, fill=green!30, line width=2pt, 
                  minimum size=1.2cm, font=\sffamily\bfseries},
    label/.style={font=\sffamily\small, color=gray!80, align=center},
    whitebox/.style={rectangle, rounded corners=12pt, draw=black, line width=2pt,
                     inner sep=15pt},
    bluebox/.style={rectangle, rounded corners=8pt, draw=blue!80, line width=2pt,
                    fill=blue!10, inner sep=10pt, minimum width=2.5cm, minimum height=4cm},
]

\node[containerbox, minimum width=4cm, minimum height=8cm] at (-4, -2.65) (leftbox) {};
\node[databox, above=1.5cm of leftbox.center] (task_init) {Task $D_{l}$};
\node[modelbox, below=1.5cm of leftbox.center] (learner_init) {An LM-Based \\ Task Learner \\ $M_{\theta+1}$};
\draw[feedbackarrow] (task_init) -- (learner_init) node[midway, sloped, label, fill=white] {Train};

\node[mlbox] at (2, 2) (generator) {Gauss-Tin \\ Generator \\ $P_{(z)}$};
\node[databox, right=3.5cm of generator] (exemplar) {Buffer $R_{t-1}$ \\ Exemplar $\hat{x}$};
\node[databox, right=1.5cm of exemplar] (task) {Task $D_{t}^{u}$};
\node[merge, below right=0.5cm and 0.25cm of exemplar] (merge) {$\oplus$};

\node[processbox, below=2cm of merge] (combined) {Task and \\ exemplar \\ $\hat{x}_{t-1} .... D_{t}^{u}$};

\node[modelbox, below=2cm of combined] (learner) {An LM-Based \\ Task Learner \\ $M_{\theta+1}$};

\coordinate (whitebox-sw) at ([xshift=-1cm, yshift=-1cm] generator.south west |- learner.south);
\coordinate (whitebox-ne) at ([xshift=1cm, yshift=1cm] task.north east |- generator.north);

\node[containerbox, fit={(whitebox-sw) (whitebox-ne)}] (mainbox) {};

\draw[mlarrow] (generator) -- (exemplar) node[midway, label, fill=white] {Generate \\ w/ Prompt \\ Assistance};
\draw[mlarrow] (exemplar) -- (merge.north west);
\draw[mlarrow] (task) -- (merge.north east);
\draw[mlarrow] (merge) -- (combined) node[midway, sloped, label, fill=white] {Merge};
\draw[feedbackarrow] (combined) -- (learner) node[midway, sloped, label, fill=white] {Train};

\draw[feedbackarrow] (combined.west) -- ++(-7.25,0) 
      node[midway, label, fill=white] {Train w/ Prompt \\ Assistance} 
     -- (generator.south);

\draw[mlarrow] (leftbox.east) -- (mainbox.west);

\node[below=0.3cm of leftbox.south, font=\sffamily\bfseries] {Initialization};
\node[below=0.3cm of mainbox.south, font=\sffamily\bfseries] {Continual Learning};

\end{tikzpicture}
}
\caption{Workflow of the Gauss-Tin Model for CL. This diagram illustrates the process flow from the Gauss-Tin Generator to the task-specific training of an LM-based Task Learner. Exemplars are generated from past task data and merged with new task data to train the model, emphasizing the minimization of catastrophic forgetting and the enhancement of task adaptability.}
\label{Gauss-Tin Framework}
\end{figure}
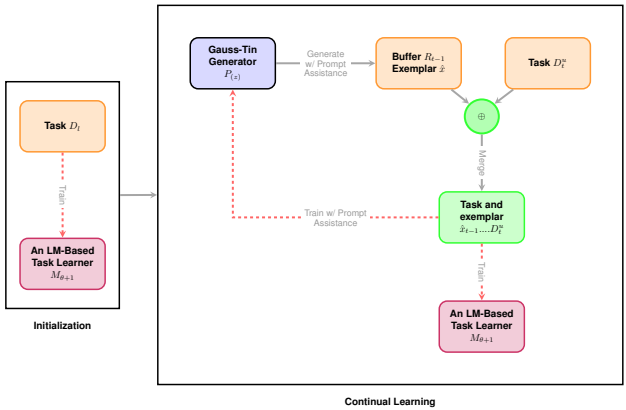

\begin{definition}
\label{def:part} Dataset partitioning. Consider a dataset $D$ partitioned into two subsets:
Labeled data $D_l : D_l=\{(x_i,y_i)\}_{i=1}^N$, where each instance $x_i$ is paired with an associated label $y_i$ and the unlabeled data $D_u : D_u =\{x_j\}_{j=1}^{N_u}$, consisting of instances without labels.
\end{definition}

The labeled subset $D_l$ is utilized during the initial training phase to enable the $M_\theta$, parameterized by $\theta$, learns useful feature representations relevant to task-incremental settings. This initial stage provides the foundation for subsequent learning processes.

\begin{definition}
\label{def:init} Initial model configuration. The model begin as $M_\theta$ and, after the initial training on $D_l$, progresses sequentially through the unlabeled tasks, evolving into $M_{\theta+1}$. 
\end{definition}

The next step is the data allocation post-intial training. The model incrementally receives subsets of the unlabeled data  $D_u^t \subset D_u$ at each stage. Each $D_u^t$ included a mixture of data related to tasks generated by the Gauss-Tin approach in $D_u^{t-1}$, alongside tasks that are new to the current stage. 

The cardinality of $D_l$ is defined manually as $k$, and for our experimental setup, we have chosen values $k = 1,5,8$ and $10$. Consequently, $D_u$ is recalculated as $D_l - k$. The overarching objective is to train $M_\theta$ to adaptively learn from $D_u^k$ at each stage,  thereby efficiently integrating new knowledge while mitigating the catastrophic forgetting of previously acquired tasks. This dual focus on forward transfer and retention underpins the theoretical and practical challenges our model seeks to address.

\subsection{Exemplar Generation}
Gauss-Tin uses a GMM module that utilizes task-specific statistics to generate high-quality exemplars. GMMs excel in modelling training sample distributions, enabling probability estimation for unknown samples and creating new samples through sampling. Their ability to identify task boundaries through density estimation makes them particularly effective for generating data in replay-based CL, highlighting their strong sampling capabilities. To enhance the effectiveness of a GMM for exemplar selection and sample generation, task-specific prompts can be leveraged to incorporate contextual information and guide the process. Since GMMs operate purely on numerical features without considering textual or contextual input, prompts provide an intermediary mechanism to focus the GMM's behaviour on task-relevant information. This integration can be applied across three key stages: \textbf{feature extraction, exemplar selection, and sample generation}. 

In the feature extraction stage, task-specific prompts are used to generate enhanced feature representations for the data. For example, a prompt like ``Cluster these samples based on their task similarity" can be processed by a pre-trained language model (e.g., BART or GPT) to create embeddings that align with the task's requirements. In our study, we employ BART to create embeddings that align with the task's requirements. Formally, given a prompt $p$, and an input sample $x$, the embedding $z$ can be represented as  $z= f_\theta(x,p)$. where $f_\theta$ is the feature extractor parameterized by $\theta$, incorporating both the input data and the task-specific prompt. These embeddings are then used as input to the GMM for clustering and density estimation.

Next step is the exemplar selection stage, task-specific prompts refine the clusters generated by the GMM. After clustering the data into $K$ Gaussian components, a prompt such as ``Select exemplars most representative of the 'Question Generation' task" can guide the identification of the most relevant samples within each cluster. This process ensures that the selected exemplars align closely with the task's objectives, prioritizing samples with higher relevance. In the initial stage, $K$ is define manually and in our experiment we set $K$ to $6$.

\begin{definition}
\label{def:exemplargen} Exemplar Generation. Given $\hat{x}$ as the exemplar, then $\hat{x} \approx \sum_{k=1}^K \pi_k \mathcal{N} (\mu_k, \sum_k)$, where $\mathcal{N}$, $\pi_k, \mu_k$ and $\sum_k$ represent the represent the Gaussian normal distribution, mixture weights, means, and covariances of the $k$-th Gaussian component, respectively. A task-specific prompt can then evaluate these samples, selecting those that best align with the desired task characteristics.
\end{definition}

\section{Experiments and Results}
This section describes the experimental setup, baseline comparisons, and details of our approach's implementation. We evaluate the effectiveness of Gauss-Tin in address catastrophic forgetting.
    \subsection{Experimental Setup}
        \subsubsection{Datasets} In this experiment, we use the Natural Instructions Version 1 dataset\cite{mishra2021cross}. Although this dataset was created to evaluate generalization performance, we can also use it to assess how well the model tackles the issue of forgetting. It consists of 6 categories, 61 tasks and each task has $\approx 20k$ instances, spanning a wide range of natural language processing (NLP) tasks. These include 13 question generation tasks (QG), 16 answer generation tasks (AG), 12 classification tasks (CF), 8 incorrect answer generation tasks (IAG), 10 minimal modification tasks (MM), and 2 verification tasks (VF). Each task provides input-output pairs along with detailed natural language instructions to guide the model in performing the task. We randomly select $j$ tasks for the evaluation split, while the remaining tasks are included in the training split. Due to resource limitations, we only utilize 10\% of the dataset which we randomly and evenly select across each category.
        
        \subsubsection{Model Specifications}  We use BART-base, a pre-trained sequence-to-sequence language model released by Huggingface \cite{lewis2019bart} and GPT-4 for generating prompts. BART (Bidirectional and Auto-Regressive Transformer) is designed for text generation and comprehension tasks. It combines a bidirectional encoder, similar to BERT, and an autoregressive decoder, similar to GPT, making it suitable for text-to-text tasks. The base version consists of 6 encoder layers and 6 decoder layers with approximately 140 million parameters. Meanwhile, GPT-4 is OpenAI's advanced LLM, designed to understand and generate highly nuanced and contextually accurate text. One of its standout capabilities is generating effective prompts or instructions for a wide range of tasks. By leveraging its extensive training on diverse datasets, GPT-4 can craft detailed and context-specific prompts that guide models or users in performing tasks efficiently \cite{achiam2023gpt}.
        
        \subsubsection{Framework and Tools} The implementation process begins by splitting the dataset $D$ into two subsets: a training set data $(t)$ and unseen tasks. The training set includes a subset of tasks such as Question Generation, Answer Generation, Classification, and Verification, while the unseen tasks are reserved for evaluation. This split ensures a diverse task distribution, allowing us to assess the model's memory across different categories. 

        To facilitate clustering, we use a task-specific prompt generated by GPT-4o. The initial prompt, \textit{``Cluster these samples based on their task similarity (e.g., Question Generation, Answer Generation, Verification, etc.),"} provides contextual guidance to group the data according to task-related features. These prompts align the input samples with the desired clustering objectives. The concatenated embeddings are then fed into a GMM for clustering. Each cluster corresponds to a specific task type category, such as Question Generation or Answer Generation, and represents the groupings of samples based on their similarity in the feature space.

        After clustering, we refine the selection process by generating a second prompt using GPT-4o: \textit{``Select the best exemplars most representative of the 'Task Type' from the clustered samples."}. The selected exemplars are then stored in a memory buffer, which serves as a repository for replay during continual learning. The memory buffer is dynamically updated as new tasks are processed, with a configurable size ranging from 10 to 50 samples per task. This ensures a balance between memory efficiency and the quality of stored exemplars. The exemplars are prioritized for their diversity and relevance to the task, enhancing the replay process and mitigating catastrophic forgetting.

        Finally, all hyperparameters are configured to optimize performance. The maximum input length is set to 1024 tokens to accommodate the BART model's capabilities. The learning rate was fixed at $5 \times10^{-5}$, and each task was trained for 3 to 5 epochs.

\begin{table*}[ht]
\caption{Performance Comparison of Continual Learning Methods on Natural Instructions Datasets}
\label{comparison_table}
\vskip 0.15in
\begin{center}
\begin{small}
\begin{sc}
\begin{tabular}{l|cccc|cccc}
\toprule
Method & \multicolumn{4}{c}{Forward Transfer ($\rightarrow$)} & \multicolumn{4}{c}{Backward Transfer ($\leftarrow$)}\\
& $D_{k1}$ & $D_{k5}$ & $D_{k8}$ & $D_{k10}$ & $D_{k1}$ & $D_{k5}$ & $D_{k8}$ & $D_{k10}$\\
\midrule
Seq-finetune (Lowerbound) & 3.28 & -3.74 & 2.90 & -0.36 & 0.04 & -0.19 & -6.48 & -9.46 \\
\midrule
\textbf{Gauss-tin (Ours)} &  \textbf{4.07} & \textbf{4.39} & \textbf{4.52} & \textbf{5.06} & $\textbf{2.21}$ & \textbf{3.33} & \textbf{5.31} & \textbf{5.99} \\
\midrule
Joint Training (Upperbound) & \multicolumn{8}{c}{6.97 $\pm$ 19.57} \\

\bottomrule
\end{tabular}
\end{sc}
\end{small}
\end{center}
\vskip -0.1in
\end{table*}

        \subsubsection{Evaluation Metrics} We use Backward Transfer (BWT) and Forward Transfer (FWT) for evaluating the performance. These metrics are specifically chosen to assess two critical aspects of CL: the model's ability to retain knowledge from previous tasks and its capability to generalize knowledge to new tasks. 
        
        \begin{definition}
            \label{def-bwt}BWT.  As our research focuses on using a replay strategy with GMM-generated exemplars to address catastrophic forgetting, therefore, BWT evaluates how well our method preserves knowledge of earlier tasks $D_1,D_2,..D_{i-1}$ when the model is incrementally trained on task $D_1$. A positive BWT score indicates effective retention of prior knowledge, while a negative score suggests forgetting. As for the method to calculate BWT, it is as follows: $
            BWT = \frac{1}{i-1}\sum_{j-1}^{i-1}(Acc(M_i,D_j) - Acc(M_j,D_j))$,  where $Acc(M_i,D_j)$ is the accuracy of the model $M_i$ (trained on $D_1,...,D_i$) on task $D_j$, and $Acc(M_j,D_j)$ is the accuracy of the model trained only on $D_j$.
        \end{definition}

        \begin{definition}
            \label{def-FWT} FWT. On the other hand, FWT evaluates how well the instructions and exemplar generation enable the model to generalize from prior tasks $D_1,D_2,..,D_{i-1}$ to future tasks $D_{1+1},D_{i+2}$,...). High FWT indicates that the model successfully utilizes previously acquired knowledge to accelerate or enhance learning on new tasks. Here is the formula for calculating the FWT:
            $FWT=\frac{1}{T-i}\sum_{j=i+1}^TAcc(M_i,D_j)$, where $Acc(M_i,D_j)$ measures the accuracy of the model $M_i$ on a future unseen task $D_j$, $i$ is the current task index up to which the model has been trained, $j$ is an index for tasks that come after $i$-th task, for which the model's performance is evaluated and $T$ is the total number of tasks.
        \end{definition}

        \subsubsection{Baseline Comparison}
        While existing research in CL often focuses on generalization, our study uniquely targets mitigating catastrophic forgetting. Therefore, we assess our model, Gauss-Tin, against traditional methods, specifically fine-tuning and joint training. Fine-tuning, where a model is sequentially adjusted to new tasks, typically leads to catastrophic forgetting, making it a pertinent baseline for demonstrating our model’s effectiveness. Joint training, training a model on all tasks simultaneously, serves as an upper bound for task retention and overall performance. Comparing Gauss-Tin against these methods highlights its advantages in scenarios where preventing catastrophic forgetting is crucial.

\section{Results and Discussion}
Our investigation into replay-based CL strategies on the Natural Instructions Dataset provides significant insights into the effectiveness of the Gauss-Tin method compared to both the traditional sequential fine-tuning approach and joint training. We are also conducting experiments to demonstrate the importance of combining prompts and GMM, as opposed to using GMM or prompts alone.

\subsection{Performance of Gauss-Tin Compared to the Baselines}
As depicted in \cref{comparison_table}, Gauss-Tin consistently outperforms the sequential fine-tuning across both forward and backward transfer metrics, while not reaching the performance levels of joint training, which serves as an upper bound. It is important to note that $k$ represents the number of initial tasks used to train the model during the initial stage, and our goal is to minimize the number of $k$ as possible. 

Gauss-Tin shows a notable improvement in forward transfer compared to sequential fine-tuning. Specifically, for tasks $D_{k5}$,$D_{k8}$, and $D_{k10}$, Gauss-Tin scores are $4.39, 4.52$ and $5.06$ respectively, demonstrating a robust ability to leverage previous knowledge in learning new tasks. This is especially evident when compared to the negative transfer observed in sequential fine-tuning for $D_{k5}$ at $-3.74$, suggesting that Gauss-Tin not only avoids the degradation of performance but also enhances learning in new contexts.

In terms of backward transfer, which assesses the model's ability to retain knowledge of previous tasks after learning new ones, Gauss-Tin significantly mitigates catastrophic forgetting. It maintains a positive score across all tasks (e.g., $2.21$ for $D_{k1}$ and $5.99$ for $D_{k10}$), in contrast to the sharp decline to $-9.46$ observed in sequential fine-tuning for $D_{k10}$. This emphasizes Gauss-Tin's effectiveness in preserving previously learned knowledge, a critical advantage in settings where models must adapt to new information without losing accuracy on older tasks.

While joint training exhibits superior performance with scores reaching up to $6.97$ and $19.57$ for forward and backward transfers respectively, it is important to note that joint training requires access to all tasks simultaneously, a condition that may not be feasible in practical scenarios where tasks arrive sequentially.

\subsection{Comparative Performance of Gauss-Tin: With Only Prompt, Only GMM, and Both}

The integration of prompts and GMM within the Gauss-Tin method for the CL framework demonstrates significant benefits. This approach significantly outperforms the use of either component alone, yielding superior results in both forward and backward transfer scenarios across various tasks, achieving an average performance score of $4.51$. Specifically, the combination of prompts and GMM leads to notable performance gains in forward transfer. The methods employing only prompts or only GMM achieved average scores of $3.145$ and $3.0325$, respectively as depicted in the \cref{comparison_table_with_without}. This indicates a more effective and efficient utilization of underlying knowledge structures. Meanwhile in backward transfer, the synergistic effect of prompts and GMM substantially enhances the retention of previous knowledge with an average performance score of $4.21$. In comparison method employing only prompts achieved average scores of $2.21$, and GMM only has an average score of $2.142$. 

The synergistic effect stems from the complementary nature of task-directed prompts and the robust data modeling provided by GMM. Prompts enhance immediate task adaptability and relevance, while GMM underpins this directed learning with a statistical foundation that captures complex data characteristics, thereby balancing the trade-offs between task specificity and learning generality. This synergy leads to a learning model that is both adaptable and robust, capable of handling complex learning environments.

These findings underscore the crucial role of integrating prompts and GMM in maximizing the effectiveness of CL framework. Future research should further delve into this synergy, exploring how different types of prompts and GMM configurations might influence outcomes across a broader spectrum of tasks and more complex datasets. Moreover, the promising results with Gauss-Tin advocate for additional studies on domain shifting within the CL framework to examine how well the method aids in generalizing across various domains. Such explorations could greatly expand Gauss-Tin’s applicability, proving its efficacy not only in knowledge retention but also in enhancing generalization to new, unseen domains. These efforts would provide deeper insights into the dynamics between specialization and generalization within the CL environment, potentially leading to the development of more robust models for dynamically changing environments.

\begin{table*}[ht]
\caption{Comparative Analysis of Methods With and Without Prompt and Gaussian Mixture Models (GMM) for Sample Generation}
\label{comparison_table_with_without}
\vskip 0.15in
\begin{center}
\begin{small}
\begin{sc}
\begin{tabular}{l|cccc|cccc}
\toprule
Method & \multicolumn{4}{c}{Forward Transfer ($\rightarrow$)} & \multicolumn{4}{c}{Backward Transfer ($\leftarrow$)}\\
& $D_{k1}$ & $D_{k5}$ & $D_{k8}$ & $D_{k10}$ & $D_{k1}$ & $D_{k5}$ & $D_{k8}$ & $D_{k10}$\\
\midrule
\textbf{Gauss-tin (Ours)} &  \textbf{4.07} & \textbf{4.39} & \textbf{4.52} & \textbf{5.06} & $\textbf{2.21}$ & \textbf{3.33} & \textbf{5.31} & \textbf{5.99} \\

\midrule
w/o Prompt & 1.88 & 3.32 & 4.41 & 2.97 & 1.74 & 2.78 & 0.83 & 3.49 \\

\midrule
w/o GMM & 2.01 & 3.89 & 2.11 & 4.12 & 1.33 & 2.14 & 1.21 & 3.89 \\

\bottomrule
\end{tabular}
\end{sc}
\end{small}
\end{center}
\vskip -0.1in
\end{table*}

\section{Conclusion}
Our research introduces the Gauss-Tin method, an innovative approach within the CL framework that effectively mitigates catastrophic forgetting in LLMs. By employing GMMs to generate exemplars and integrating a strategic replay mechanism, Gauss-Tin substantially enhances both forward and backward transfer capabilities compared to conventional sequential fine-tuning and joint training methods. The approach not only maintains knowledge across sequential tasks but also adapts proficiently to new tasks, significantly reducing the loss of prior learning. These findings underscore the potential of Gauss-Tin to improve model adaptability and efficiency, suggesting further exploration into its impact on generalization across varied domains could extend its applicability and effectiveness in dynamic learning environments.

\nocite{langley00}

\bibliography{gauss_tin_paper}
\bibliographystyle{icml2025}

\end{document}